\documentclass[sigconf]{acmart}
\renewcommand\footnotetextcopyrightpermission[1]{} 
\usepackage{algorithm}
\usepackage{algorithmic}
\usepackage{amsmath}
\AtBeginDocument{%
  \providecommand\BibTeX{{%
    \normalfont B\kern-0.5em{\scshape i\kern-0.25em b}\kern-0.8em\TeX}}}

\begin{document}

\title{Generator and Critic: A Deep Reinforcement Learning Approach for Slate Re-ranking in E-commerce}

\author{Jianxiong Wei}
\authornote{Both authors contributed equally to this research.}
\email{jianxiong.wjx@alibaba-inc.com}
\author{Anxiang Zeng}
\authornotemark[1]
\email{renzhong@taobao.com}
\affiliation{%
  \institution{Alibaba Group}
}

\author{Yueqiu Wu}
\email{yueqiu.wu@lazada.com}
\affiliation{%
  \institution{Alibaba Group}}

\author{Peng Guo}
\email{peng.guop1@alibaba-inc.com}
\affiliation{%
  \institution{Alibaba Group}}

\author{Qingsong Hua}
\email{qingsong.huaqs@taobao.com}
\affiliation{%
  \institution{Alibaba Group}}

\author{Qingpeng Cai}
\email{qingpeng.cqp@alibaba-inc.com}
\affiliation{%
  \institution{Alibaba Group}
}

\renewcommand{\shortauthors}{Wei and Zeng, et al.}

\begin{abstract}
The slate re-ranking problem considers the mutual influences be- tween items to improve user satisfaction in e-commerce, compared with the point-wise ranking. Previous works either directly rank items by an end to end model, or rank items by a score function that trades-off the point-wise score and the diversity between items. However, there are two main existing challenges that are not well studied: (1) the evaluation of the slate is hard due to the complex mutual influences between items of one slate; (2) even given the optimal evaluation, searching the optimal slate is challenging as the action space is exponentially large. In this paper, we present a novel Generator and Critic slate re-ranking approach, where the Critic evaluates the slate and the Generator ranks the items by the reinforcement learning approach. We propose a Full Slate Critic (FSC) model that considers the real impressed items and avoids the ``impressed bias'' of existing models. For the Generator, to tackle the problem of large action space, we propose a new exploration reinforcement learning algorithm, called PPO-Exploration. Experimental results show that the FSC model significantly outperforms the state of the art slate evaluation methods, and the PPO-Exploration algorithm outperforms the existing reinforcement learning methods substantially. The Generator and Critic approach improves both the slate efficiency(4\% gmv and 5\% number of orders) and diversity in live experiments on one of the largest e-commerce websites in the world.
\end{abstract}




\maketitle
\thispagestyle{empty}
\pagestyle{plain} 

\section{Introduction}
\begin{figure}[H] 
\centering
\includegraphics[width=0.4\linewidth]{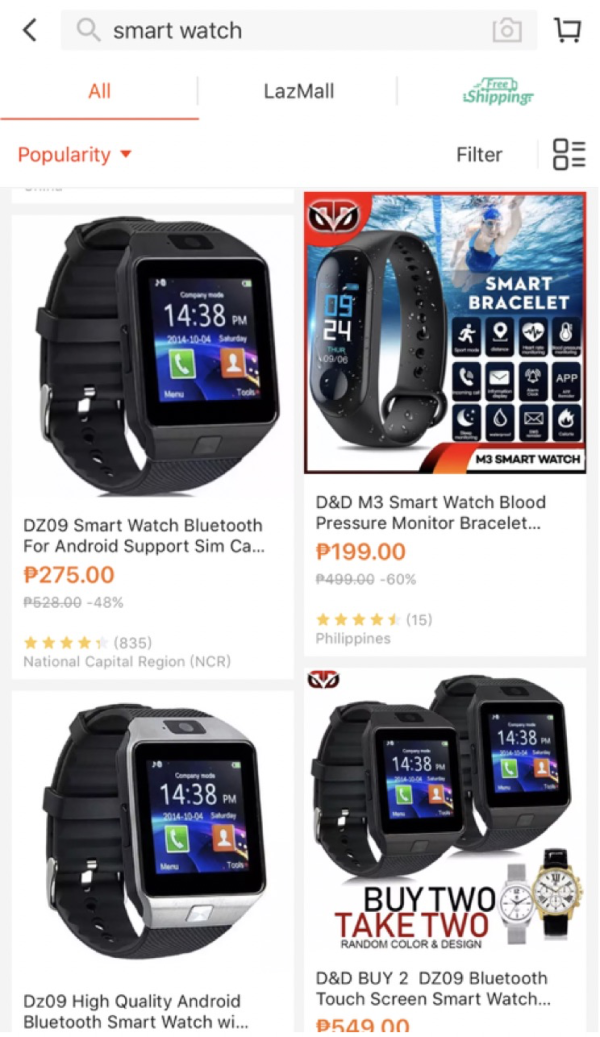}

\caption{The return list when searching ``smart watch''}
\label{fig:smart_watch}
\end{figure}

In a typical e-commerce website, when a user searches a keyword, the website returns a list of items to the user by the ranking algorithm. This process is usually done by predicting the scores of user-item pairs and sorting items based on the point-wise scores. However, the point-wise ranking algorithm does not consider the mutual influences between items in one page, and returns similar items with the highest scores. Here is an example in Figure \ref{fig:smart_watch}: a buyer searches the “smart watch" in one e-commerce app, the app returns similar watches, which will decrease user satisfaction and the efficiency of the list.

To improve the diversity of the list, a series of research works, MMR, IA-Select, xQuAD and DUM \cite{agrawal2009diversifying,ashkan2015optimal,carbonell1998use,santos2010exploiting} are proposed to rank items by weighted functions that trade-off the user-item scores and the diversities of items. However, these methods ignore the impact of diversity on the efficiency of the list. Wilhelm et.al \cite{wilhelm2018practical} present a deep determinantal point process model to unify the user-item scores and the distances of items, and devise a practical personalized approach. Nevertheless, the determinantal point process uses separate structures to model the user-item scores and the diversities of items, which limits the representation ability of the user interest over the slate.

\cite{ai2018learning, ai2019learning, zhuang2018globally}
propose models that input the n point-wise ranked items, and output the re-fined scores, by adding intra-item features to the model. This approach selects top-k items from n refined scores, and is commonly called ``Slate Re-ranking'' or ``Post Ranking''($n\geq k$). \cite{pei2019personalized} employ the efficient Transformer structure and introduce the personalized user encoding to the structure, compared with the RNN structure in two classical re-ranking methods, DLCM \cite{ai2018learning} and GlobalRank \cite{zhuang2018globally}. However, the main concern of this approach is that only $k$ items are shown (impressed) to the user in reality. Thus these models can not reflect the true preference of the user over the slate, which causes the ``impressed bias''.

Indeed, the slate re-ranking problem presents two main challenges:

(1) The evaluation of the slate is difficult, as it is crucial for a model to tackle the ``position bias'', the ``impressed bias'', and the mutual influence of items, where the ``position bias'' means the impact of the positions of items on the preference of the user.

(2) Given an evaluation model for any top-$k$ slate, the number of possible choices of top-$k$ slates is extremely huge.

To solve these two challenges, we propose a novel Generator and Critic slate re-ranking approach, shown in Figure 2. The Full Slate Critic module evaluates any candidate top-$k$ list. The Generator learns the optimal top-$k$ list by reinforcement learning algorithms \cite{sutton2018reinforcement}. That is, the Generator module is the reinforcement learning agent and the Critic module is the environment. The environment is defined as follows: At each episode(a user and a query come), the action of the Generator at each step is to pick an item from $n$ candidate items, and the Critic returns the evaluation score(reward) of the whole slate after the episode ends, i.e., $k$ items have been selected. At each step, the Generator takes a reinforcement learning policy that inputs the state(the user, the query, the feature of $n$ items, and selected items), outputs an action(the new item to be added to the list).

For the Critic module, to tackle the exact evaluation challenge of the slate, we present a Full Slate Critic(FSC) model. The FSC model inputs features of $k$ items, rather than $n$ total items, which avoids the ``impressed bias''. The model consists of four components: Su Attention Net, Pair Influence Net, Bi-GRU Net and Feature Compare Net. The Su Attention Net encodes the real-time impressed items, click and pay behaviors of the user in the same search session, to capture the real-time interest of the user. The Pair Influence Net calculates the pair-wise influences of other items on each item, and aggregates the influences by the Attention method \cite{wang2017residual}, which models the mutual influences between items. The Bi-GRU Net uses Bidirectional-GRU structure \cite{schuster1997bidirectional} to capture the influence of the nearby items over each item, and is designed for reducing ``position bias''. The Feature Compare Net compares the value of each item with other items, and represents the impact of the difference of feature values on the user interest.

For the Generator module, we embrace the recent advancements of combining deep reinforcement learning with e-commerce \cite{cai2018reinforcementa, cai2018reinforcementb, chen2019top, chen2018stabilizing, chen2018generative, hu2018reinforcement, zhao2018deep, zheng2018drn}. Chen et.al \cite{chen2019top} proposes a top-$k$ off-policy correction method that takes the reinforcement learning(RL) policy to generate the top-$k$ list, and learns the policy from real rewards. However, this approach is not practical as it costs millions of samples to train a good policy in a model-free reinforcement learning style, and the e-commerce site can not afford the exploration risk. As model-based RL methods are significantly more sample efficient than model-free RL methods \cite{deisenroth2011pilco, levine2016end}, we train the RL policy with model-based samples, which are generated by the RL policy and evaluated by the Critic module. We present a new RL method, called PPO-Exploration, which builds on the state of the art reinforcement learning method, proximal policy optimization(PPO)\cite{schulman2017proximal}. PPO-Exploration drives the policy to generate diverse items, by adding the diversity score to the reward function at each step. The diversity score is defined as the distance between the current picked item over selected items and serves as the exploration bonus \cite{bellemare2016unifying}.

To summarize, the main contribution of this paper is as follows:

1. We present a new Generator and Critic slate re-ranking approach to tackle the two challenges: the exact evaluation of slates and the efficient generation of optimal slate.

2. We design a Full Slate Critic(FSC) model avoiding the ``im pressed bias'' that exists in current methods, and better represents the mutual influences between items. Experimental results show that the FSC model outperforms both the point-wise Learning to Rank (LTR) and the DLCM method substantially. We show that the FSC model is sufficiently correct to evaluate any slate generation policy.

3. We propose a new model-based RL method for generating slates efficiently, called PPO-Exploration, which encourages the policy to pick diverse items in the slate generation. Experimental results show that the PPO-Exploration performs better than the reinforce algorithm and the PPO algorithm significantly. We also validate the effectiveness of the model-based method comparing with the model-free method that trains the RL policy with real rewards.

4. The Generator and Critic approach has been successfully ap plied in one of the largest e-commerce websites in the world, and improves 5\% number of orders and 4\% gmv during the A/B test. The diversity of slates is also improved.

\subsection{Related Work}
Deep Reinforcement Learning(DRL) recently achieves cutting-edge progress in Atari games \cite{mnih2015human} and continuous control tasks \cite{lillicrap2015continuous}. Motivated by the success of deep reinforcement learning on these tasks, applying DRL on the recommendation and searching task has been a hot research topic recently 
\cite{bai2019model, chen2019large, chen2019top, chen2018generative, gong2019exact, shi2019virtual, takanobu2019aggregating, zhang2019text, zou2019reinforcement}. The most representative work in this research line is \cite{hu2018reinforcement}, where Hu et.al apply reinforcement learning to the Learning to Rank task in e-commerce.

The main difference between our work and previous works is that we do not solve the MDP where the agent is the recommendation system, at each state the action of the agent is to recommend $k$ items to the user. In our MDP, there are one new user and one query at each initial state of each episode, the action is to pick one item from $n$ original LTR ranked items at each step. To the best of our knowledge, we are one of the first to applying reinforcement learning on the slate re-ranking problem. We also build a more precise evaluation model of the slate, which enables efficient model- based reinforcement learning.

Different from slate re-ranking works that focus on the refined scores of items, Jiang et.al \cite{jiang2018beyond} propose List-CVAE to pick $k$ items by item embedding. List-CVAE uses a deep generative model to learn $k$ desired embedding, and picks $k$ items closest to each desired embedding from the original $n$ items. However, this approach can not guarantee the efficiency of the output list as there may not exist $k$ items that are close enough to the desired embedding.

\section{Preliminary}
In this section we introduce the basic concepts of slate re-ranking in e-commerce, and reinforcement learning.

\subsection{Slate Re-ranking}
When a user $u_i$ searches a query $q_i$, the ranking system returns $n$ items $x_i=(x_{i1},...,x_{in})$ sorted by n user-item pair scores $$(s(q_i,u_i,x_{i1}),...,s(q_i,u_i,x_{in})).$$ Let $f(q_i,u_i,x_i)=L_k$ denotes the slate re-ranking function. The slate re-ranking function outputs a top-k list $L_k=(x_{ij_1},...,x_{ij_k})$, given the input of the query, the user feature, and n items.

Given an estimation function $p(q_i,u_i,L_k)$, the score of top-k list $L_k$ with the query $q_i$ and the user $u_i$, the objective of slate re-ranking is to find a slate re-ranking function $f$ that maximizes the total user satisfaction over m queries, $E(f)=\sum_{i=1}^{m}p(q_i,u_i,f(q_i,u_i,x_i))$.

\subsection{Reinforcement Learning}
Reinforcement learning focus on solving the Markov Decision Process problem \cite{sutton2018reinforcement}. In a Markov Decision Problem(MDP), the agent observes a state $s$ in the state space $\mathcal{S}$, plays an action $a$ in the action space $\mathcal{A}$, and gets the immediate reward $r(s,a)$ from the environment. After the agent plays the action, the environment changes the state by the distribution, $p(s'|s,a)$, which represents the probability of the next state $s'$ given the current state $s$ and action $a$. The policy is defined as $\pi_{\theta}(a|s)$, that outputs an action distribution for any state $s$,
and is parameterized by $\theta$. The objective of the agent is to find $\theta$
maximizing the expected discounted long-term rewards of the policy: $J(\pi_{\theta})=E[\sum_{t=0}^{\infty}{\gamma}^{t}r(s_t,a_t)|s_0,\pi_{\theta}]$, where $s_0$ denote the initial state, and $\gamma(0\leq \gamma<1)$ denotes the discounted
factor.

The well-known DQN algorithm \cite{mnih2015human} applies the maximum operator on the action space, which is not suitable to the slate re-ranking problem with large action space. Thus in this paper we focus on stochastic reinforcement learning algorithms. The Reinforce algorithm \cite{williams1992simple} is proposed to optimize a stochastic policy by the gradient of the objective over the policy and is applied in \cite{chen2019top}: $\nabla_{\theta}J(\pi_{\theta})=E_{s_t,a_t\sim {\pi}_{\theta}}[R(s_t,a_t)\nabla_{\theta}log{\pi}_{\theta}(a_t|s_t)],$ where $R(s_t,a_t)=\sum_{t'=t}^{\infty}{\gamma}^{t'-t}r(s_{t'},a_{t'})$ denotes the discounted rewards from the $t$-th step. However, the Reinforce algorithm suffers from the high variance problem as it directly uses the Monte-Carlo sampling method to estimate the long-term return. Thus actor-critic algorithms such as a2c \cite{mnih2016asynchronous} chooses to update the policy by 
\begin{equation}
\label{policy_gradient}
E_{s_t,a_t\sim \pi_{\theta}}[A_t\nabla_{\theta}log\pi_{\theta}(a_t|s_t)],    
\end{equation}

where $A_t$ is the advantage function, and is defined as the difference between the action-value function $Q(s_t,a_t)$ and the value function $V(s_t)$. But the policy gradient of Eq.(\ref{policy_gradient}) may incur large update of the policy, that is, the performance of the policy is unstable during the training. Schulman et.al \cite{schulman2017proximal} propose the proximal policy optimization(PPO) algorithm that optimizes a clipped surrogate objective to stabilize the policy update:

\begin{equation}
L^{clip}(\theta,{\theta}^{'})=E_{s_t,a_t\sim \pi_{\theta}}[min\{r_t(\theta,{\theta}^{'})A_t, clip(r_t(\theta,{\theta}^{'}),1-\epsilon,1+\epsilon)A_t\}],
\end{equation}

where $r_t(\theta,{\theta}^{'})=\frac{\pi_{\theta^{'}}(a_t|s_t)}{\pi_{\theta}(a_t|s_t)}.$ The PPO algorithm enables efficient computation compared with the trust-region policy optimization method \cite{schulman2015trust}, and achieves state of the art performance on standard control tasks. In this paper we adapt the principle of the PPO algorithm to train the slate generation policy.

\section{THE GCR MODEL: THE GENERATOR AND CRITIC RANKING APPROACH}

In this section we introduce the Generator and Critic ranking framework, called the GCR model. The GCR model works as follows, shown in Figure \ref{fig:gcr}:

\begin{figure}[H] 
\centering
\includegraphics[width=1.0\linewidth]{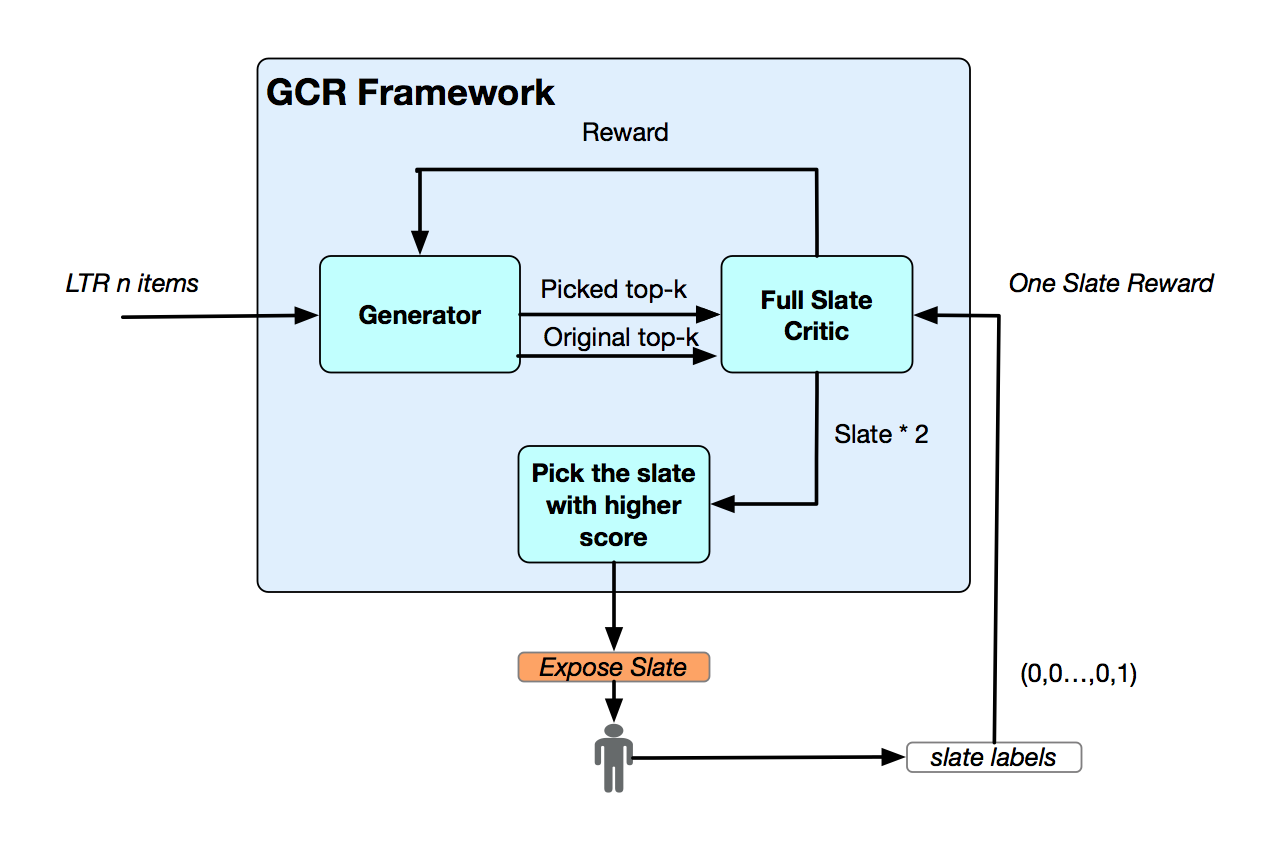}

\caption{The GCR framework}
\label{fig:gcr}
\end{figure}

\begin{itemize}
    \item When a user query comes to the system, the LTR model picks $n$ items to the GCR model.
    \item The Generator module picks top-$k$ items from $n$ items by the reinforcement learning policy.
    \item The Critic module evaluates both the picked top-$k$ slate and the original top-$k$ slate.
    \item The GCR model chooses the slate with a larger score from two candidate slates: the original slate and the re-ranked slate, and exposes the slate to the user.

\end{itemize}

The Critic module is trained with real user feedback on the slate, and serves as the environment. The Generator module, i.e., the agent, generates slates, receives the evaluation scores of slates from the Critic, and updates the reinforcement learning policy based on these evaluation scores.

\section{THE CRITIC MODULE}
In this section we present the Full Slate Critic model that avoids the ``impressed bias'', reduces the ``position bias'', and handles the mutual influences of items precisely.

\subsection{The Bias of Slate Evaluation}
First of all, we discuss the limitation of previous works on the slate evaluation. We claim that it is crucial for a slate evaluation model to consider exact items that are impressed to the users in reality.

\begin{figure}
\centering
\includegraphics[width=1.0\linewidth]{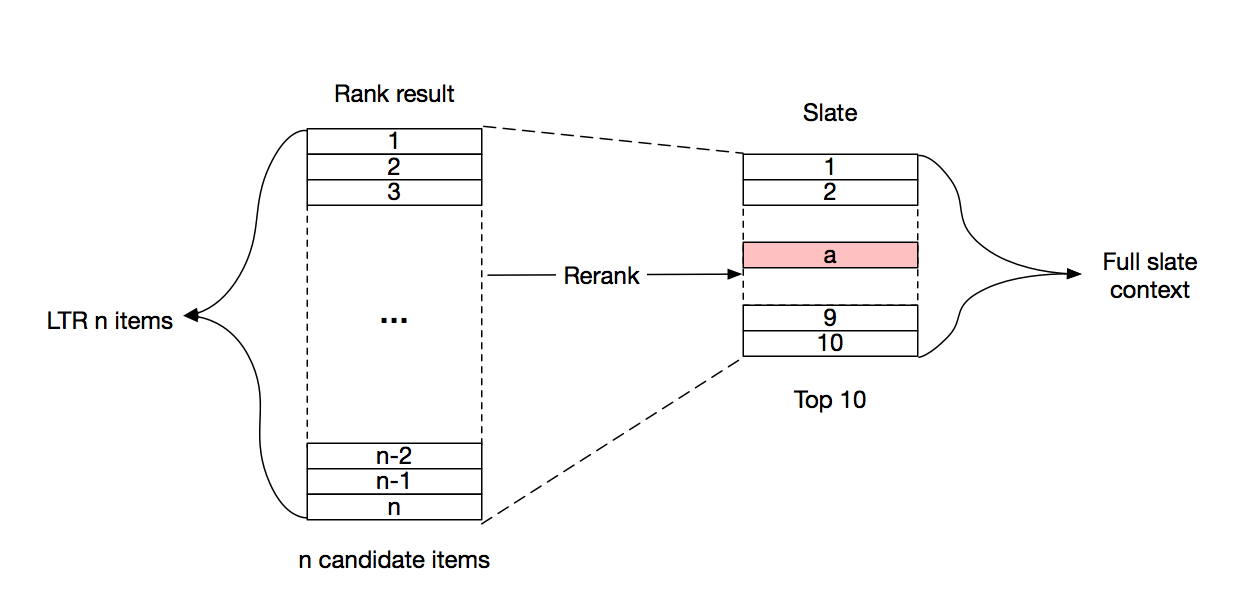}

\caption{The bias of slate evaluation}
\label{fig:slate_bias}
\end{figure}

As shown in Figure \ref{fig:slate_bias}, given $n$ candidate items, the slate re- ranking process outputs $k$ items to the user. In most e-commerce websites, $k$ is equal to 10. \cite{ai2018learning, ai2019learning, zhuang2018globally} propose models that input the $n$ items and outputs $n$ refined scores by considering features between items. But only top-$k$ items are shown to the user, and the evaluation model that builds upon $n$ items can not exactly reflect the true user preference over any real impressed slate. That is, there exists the ``impressed bias''. Also, picking top-$k$ items from refined scores limits the search space of the slate generation.

\subsection{The Full Slate Critic Model}
Now we introduce the critic module, named as the Full Slate Critic(FSC) model. The FSC model takes user behavior sequences $u_i$ one search session, features of $k$ items, the query $q_i$ as input, outputs the predicted conversion probabilities of $k$ items, denoted by $p_i=(p_{i1},...,p_{ia},...,p_{ik}).$
For the ease of the simplicity, we let $p_{ia}$
denote the predicted conversion probability of the item $a$.

As the FSC model already considers the influence of other items over each item when predicting the score of each item, we assume that the conversion probabilities of items are independent. Then the conversion probability of the whole slate (at least one item in the slate is purchased) predicted by the model is:

\begin{equation}
\label{obj}
p_i(q_i,u_i,L_k)=1-\prod_{a=1}^{k}(1-p_{ia}).
\end{equation}
The objective of the model is to minimize the cross entropy between the predicted scores of items $p_i$
and the true labels of item $t_i$, $L(\theta)=\sum_{i=1}^{m} crossentropy(p_i,t_i).$

\begin{figure}
\centering
\includegraphics[width=1.0\linewidth]{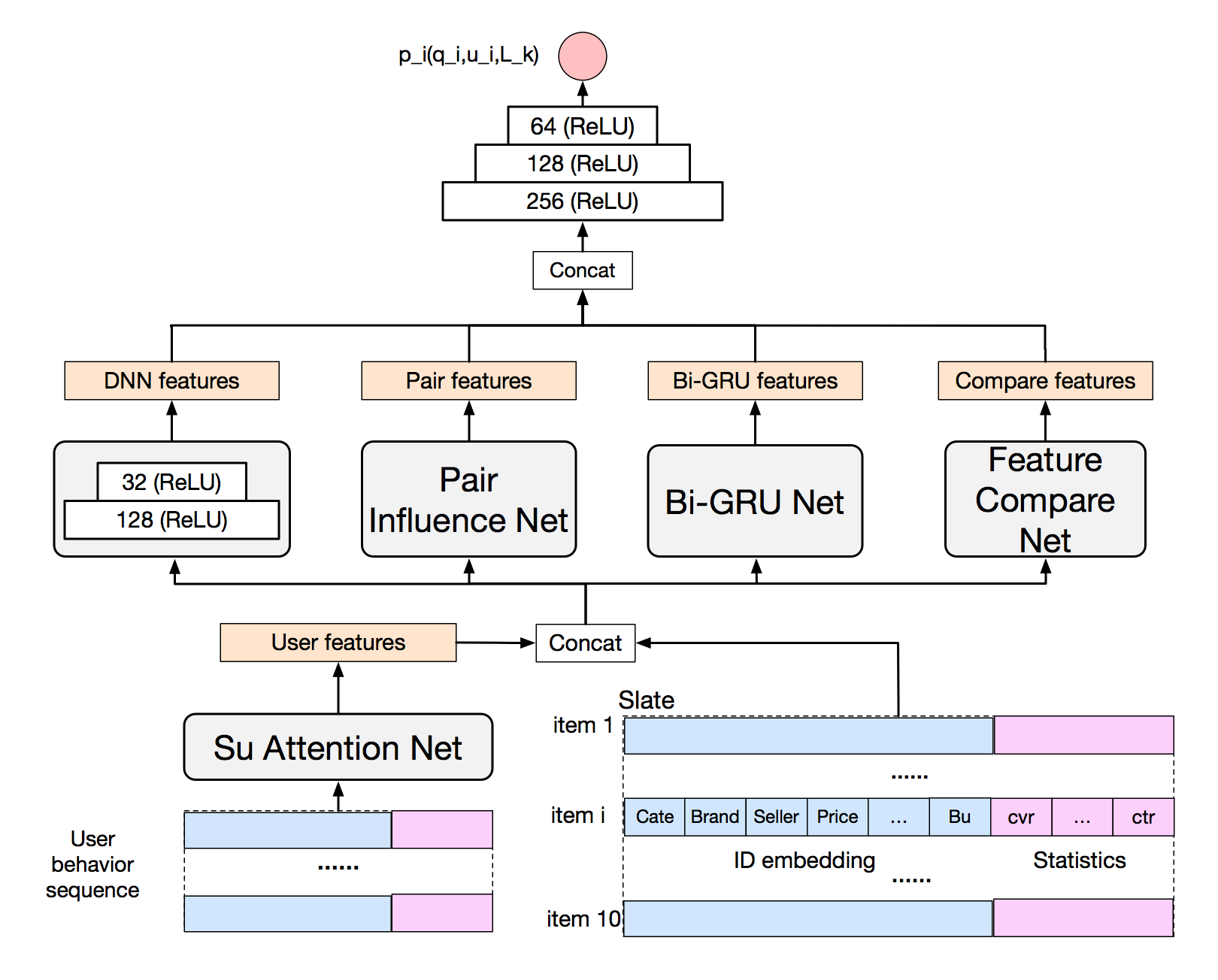}

\caption{The Full Slate Critic Model}
\label{fig:fsc}
\end{figure}

For the architecture of the Full Slate Critic(FSC) model, there are four parts: Su Attention Net, Pair-Influence Net, Bi-GRU Net and Feature Compare Net, shown in Figure \ref{fig:fsc}.

{\bf Su Attention Net:} The samples are firstly processed by the model, where the static ID features (Item ID, Category ID, Brand ID, Seller ID) and statistics (cvr,ctr) of 10 items are concatenated with the processed user real-time behavior sequences by the Su Attention Net.

\begin{figure}
\centering
\includegraphics[width=1.0\linewidth]{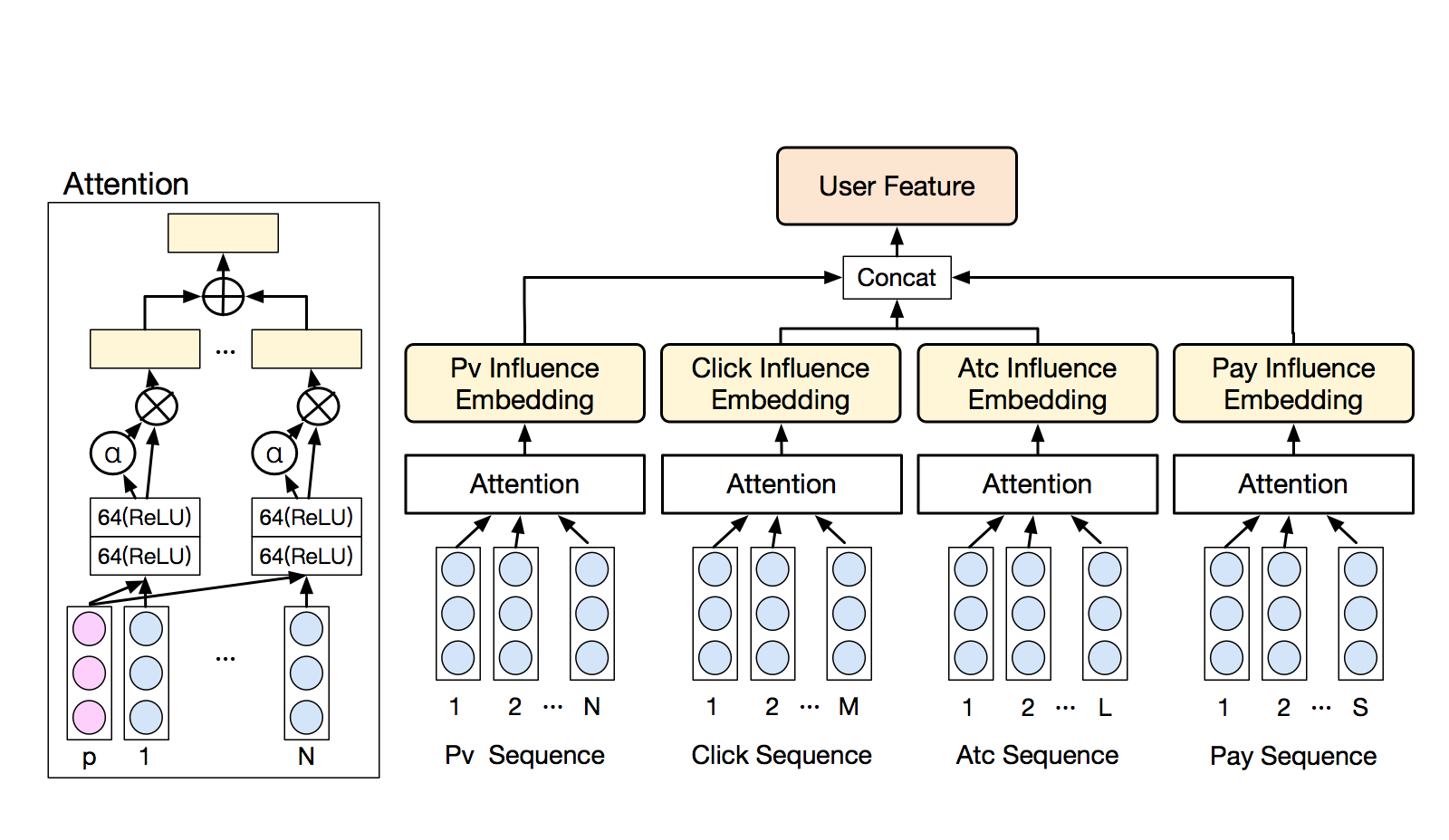}

\caption{Su Attention Net}
\label{fig:san}
\end{figure}

The detail of the Su Attention Net is shown in Figure \ref{fig:san}. The impact of the products exposed in the same search session and the real-time behavior of the user on the current product is aggregated using the attention technique. We separately process impression(Pv), click, add to cart(Atc), and purchasing(pv) item feature lists to get Pv influence embedding, Click influence embedding, Atc influence embedding, and Pay influence embedding respectively. Then these embeddings are concatenated with 10 item features for use by subsequent networks.

\begin{figure}
\centering
\includegraphics[width=1.0\linewidth]{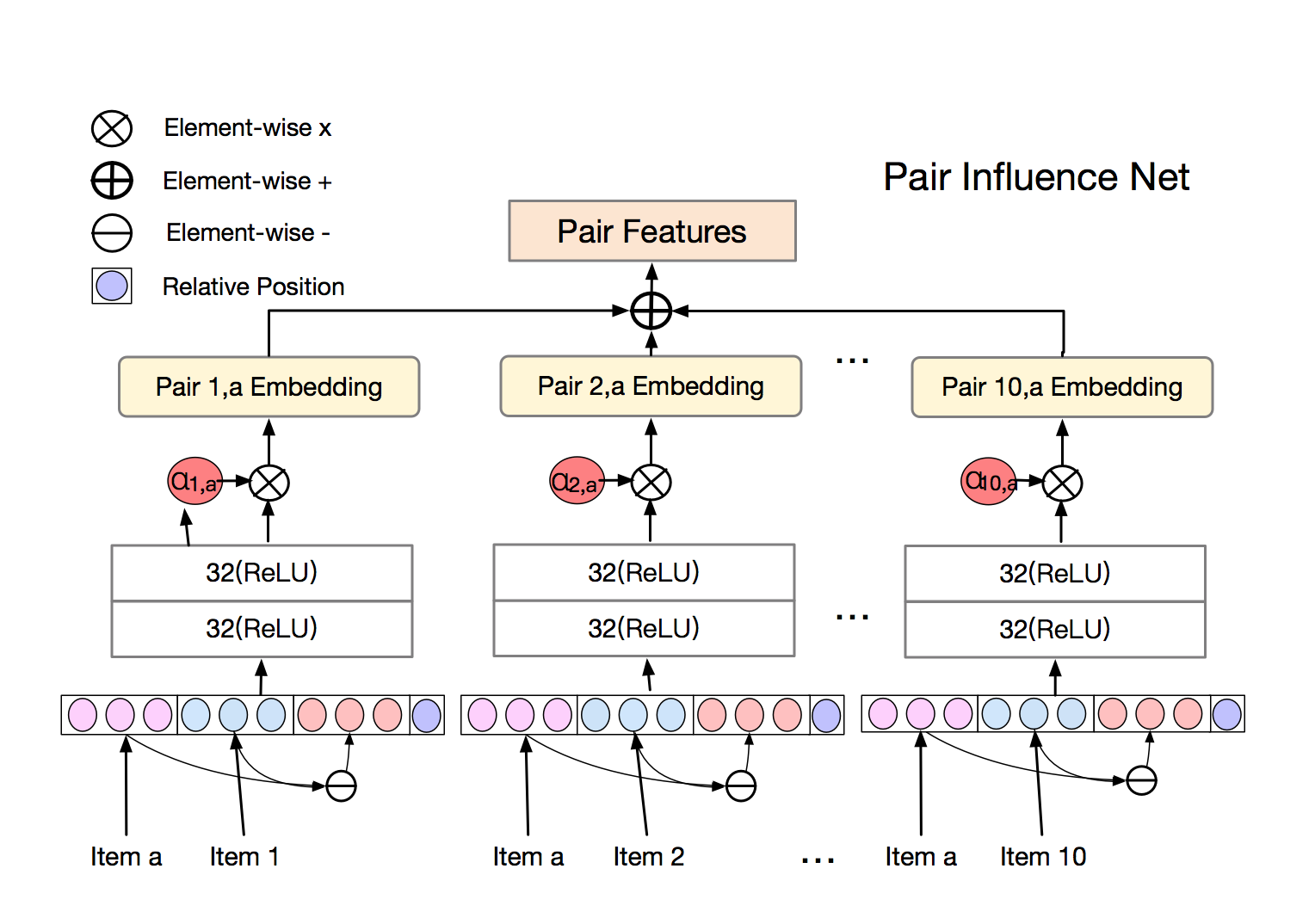}

\caption{Pair Influence Net}
\label{fig:pin}
\end{figure}

{\bf Pair Influence Net:} We use the Pair Influence Net to model mutual influences between items inside one slate. For example, if the price of one item among the 10 same items is higher than the other 9 items, the probability of this item being purchased will decrease. For each item $a$, we firstly obtain the pair-wise influence of other items over this item, which models the impact of both features and relative positions. Let $V_{ja}$  denote the influence of item
$j$ over the item $a$. The total influence of other items over the item $a$, $I_a$ is represented by aggregating $k$
pair-wise influences with the attention technique:

\begin{equation}
\begin{split}
&I_a=\sum_{j=1}^{k}{\alpha}_{ja}V_{ja},\\ &{\alpha}_{ja}=softmax_{\beta} (w_jV_{ja}+b_j)=\frac{e^{\beta(w_jV_{ja}+b_j)}}{\sum_{j=1}^{k}e^{\beta(w_jV_{ja}+b_j)}},
\end{split}
\label{attention}
\end{equation}

where $\alpha_{ja}$ denote the weight of the pair $(j,a)$, and is obtained by the softmax distribution. The detailed structure of the Pair Influence Net is referred to Figure \ref{fig:pin}.

\begin{figure}
\centering
\includegraphics[width=1.0\linewidth]{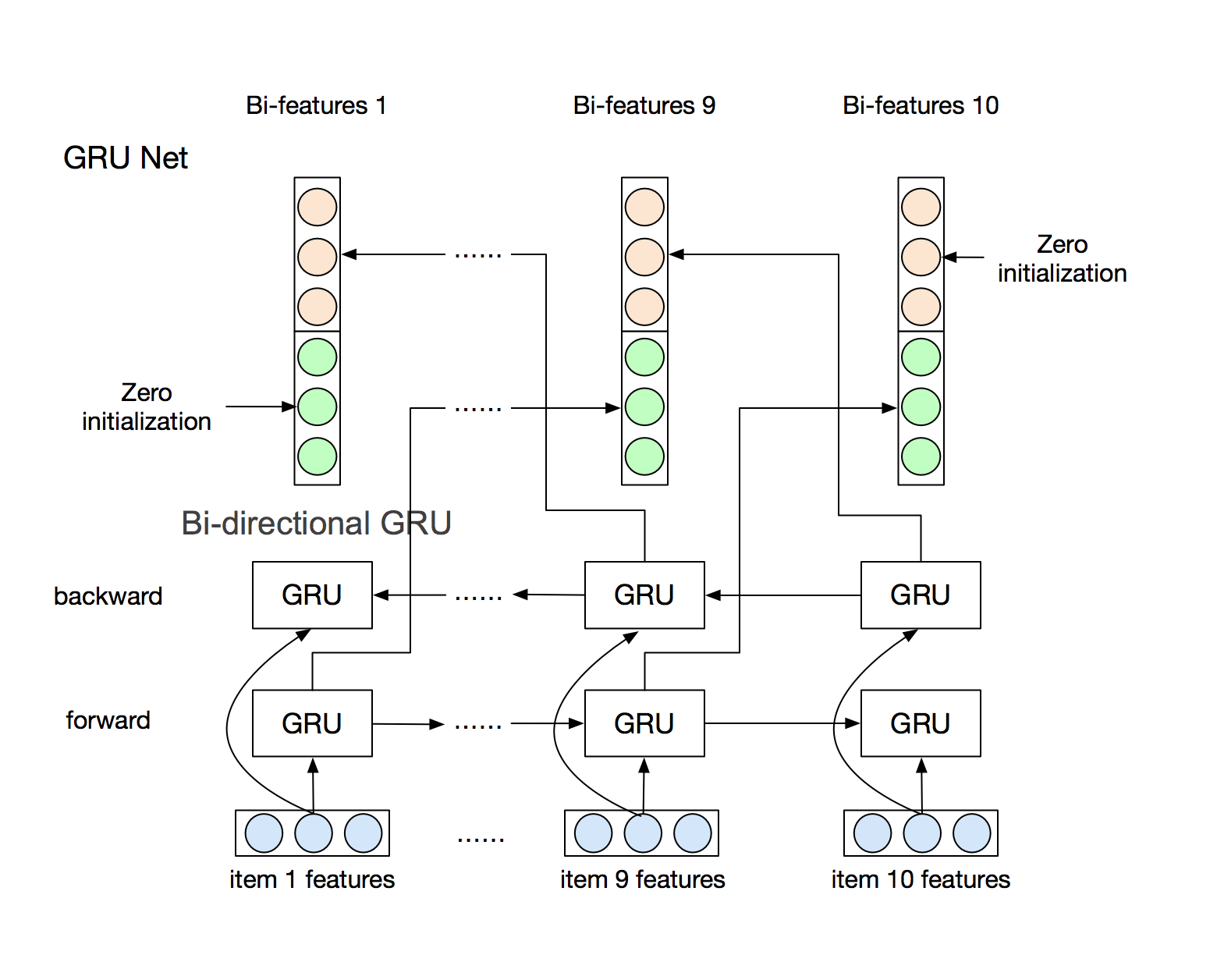}

\caption{Bi-GRU Net}
\label{fig:bi-gru_net}
\end{figure}

{\bf Bi-GRU Net:} The position bias of items is handled by the Bi-GRU Net, which uses Bidirectional Gate Recurrent Unit(GRU) to model the influence of nearby items over each item $a$, as shown in Figure \ref{fig:bi-gru_net}. The position influence of each item $a$ is the concatenation of the results of the $a-1$-th forward GRU and $a+1$-th backward GRU.

\begin{figure}
\centering
\includegraphics[width=1.0\linewidth]{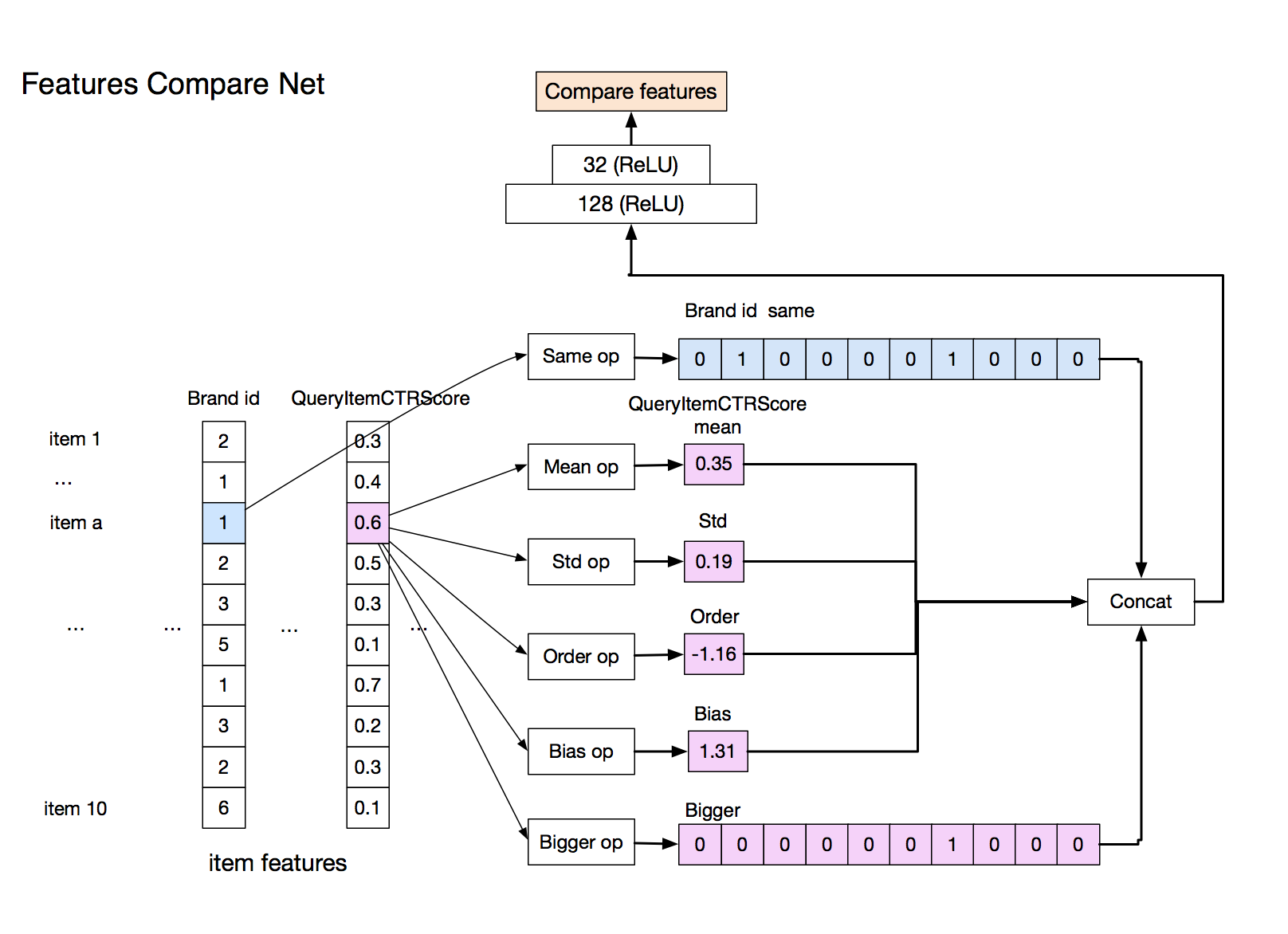}

\caption{Feature Compare Net}
\label{fig:fcn}
\end{figure}

{\bf Feature Compare Net:} The features of items are from two categories, discrete ID-type features (item ID, category ID, brand ID, etc.) and real-value features (statistics, user preference scores, etc.). The Feature Compare Net in Figure \ref{fig:fcn} takes these feature values as input, output the comparison result of the values of other items for each item and each feature. The Feature Compare Net enables more efficient structures for slate evaluation, as it directly encodes the difference of items in the dimension of feature values.

\section{THE GENERATOR MODULE}
In this section, we adapt the reinforcement learning method to search the optimal slate given the above Full Slate Critic model. We firstly formally define the slate re-ranking MDP, discuss three main challenges of this MDP, and propose the PPO-Exploration algorithm that tackles these challenges.

\subsection{The Slate Re-ranking MDP}
Recall the definition of the slate re-rank function defined in Section 2, $f(q_i,u_i,x_i)=L_k$. We choose to generate the slate by picking up items one-by-one from the candidate list $x_i$. Formally, the slate re-ranking MDP is defined as follows:

{\bf State:} There are $k$ steps in one episode, and the state at each step $t$, $s_t$ is defined as a tuple $(q_i,u_i,x_i,g_{it})$, where $g_{it}$ denotes the set of selected items before the $t-$th step.

{\bf Initial state:} At the first step of each episode, a query $q_i$ and user $u_i$ is sampled from a fixed distribution, then the LTR model generates $n$   candidate items. The initial state $s_1$ is a tuple $(q_i,u_i,x_i,g_{i1})$, where $g_{i1}$ is empty in the initial state.

{\bf Action:} The action of the policy at each state $s_t$
is the index of a candidate item, denoted by $a_t\in [1,...,n]$.

{\bf State transition:} The state transition is defined as
\begin{equation}
g_{it+1}=g_{it}\cup a_t, s_{t+1}=(q_i,u_i,x_i,g_{it+1}).
\end{equation}

The set $g_{it+1}$ represents the set of items that are selected before the $t+1$-th step.

{\bf Reward:} The objective of the generator is to maximize the total conversion probability of the total slate, Eq.(\ref{obj}). The immediate reward $r(s_t,a_t)$ is 0 
before the whole slate is generated, and is $p_i(q_i,u_i,g_{ik})$ at the last step of the episode.

{\bf Done:} Each episode ends after $k$ steps.

\subsection{Challenges of the Slate Re-ranking MDP}
There are three challenges to solve the slate re-ranking MDP:

{\bf State representation:} Modeling the influence of selected items over the remaining items is critical for the RL policy to pick up diverse items in subsequent steps.

{\bf Sparse reward:} By the definition of the slate re-ranking MDP, the reward function is sparse except for the last step, which renders RL algorithms from learning better slates. That is, to learn better policies, a more appropriate reward function is required.

{\bf Efficient exploration:} The number of possible slates is $C(n,k)$ is extremely large. Although the PPO algorithm enables the stable policy update, it tends to stuck in the local optimum and can not explore new slates. Thus it is vital to improve the exploration ability of the RL algorithms.

\subsection{The PPO-Exploration Algorithm}
Now we present our design to tackle these three challenges.

\begin{figure}
\centering
\includegraphics[width=1.0\linewidth]{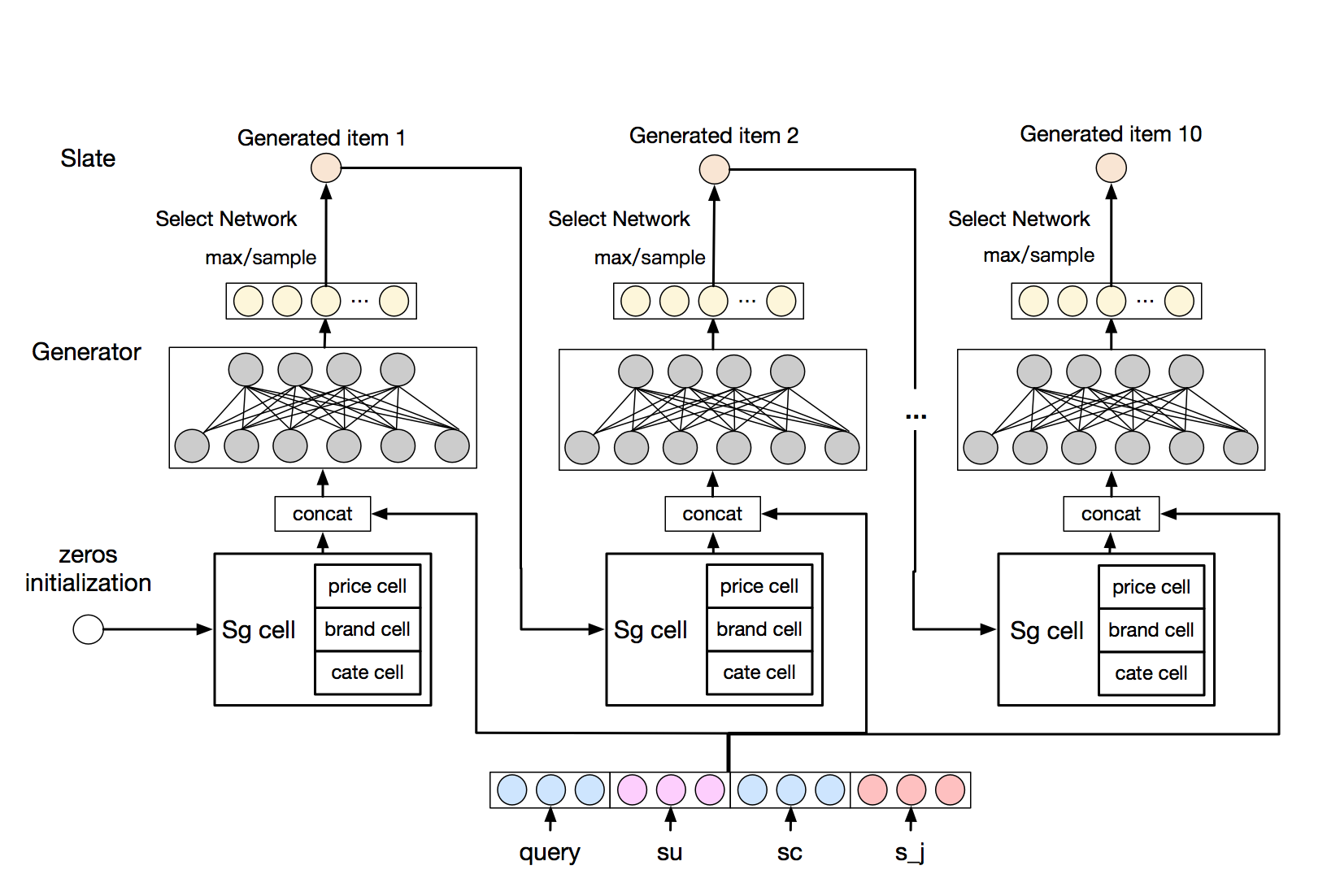}

\caption{ The Policy Network}
\label{fig:policy}
\end{figure}

{\bf State representation:} The design of the policy network, as shown in Figure \ref{fig:policy}. By definition, the state at each time step $t$ is $s_t=(q_i,u_i,x_i,g_{it})$. The policy
processes the state and inputs the vector ((query, su, sc, $s_1$, $sg_{t1}$), ..., (query, su, sc, $s_n$, $sg_{tn}$)), where su is the the status of the user behavior sequences, sc denotes the features of $n$ candidate items, $s_j$
denotes the feature of the candidate item $j$
and $sg_t$ represents the information of selected items generated by the Sg cell. Then the policy outputs weights $w=(w_1,...,w_n)$ of 
candidate items, and samples an item from the softmax distribution,
i.e., $\pi_{\theta}(a_t=j|s_t)=\frac{e^{\beta w_j}w_j}{\sum_{j=1}^{k}e^{\beta w_j}w_j}$. Note that during the training of the RL policy, the softmax distribution is used to sample actions and improve the exploration ability, while the policy selects the item with the maximum score in testing. 

\begin{figure}
\centering
\includegraphics[width=1.0\linewidth]{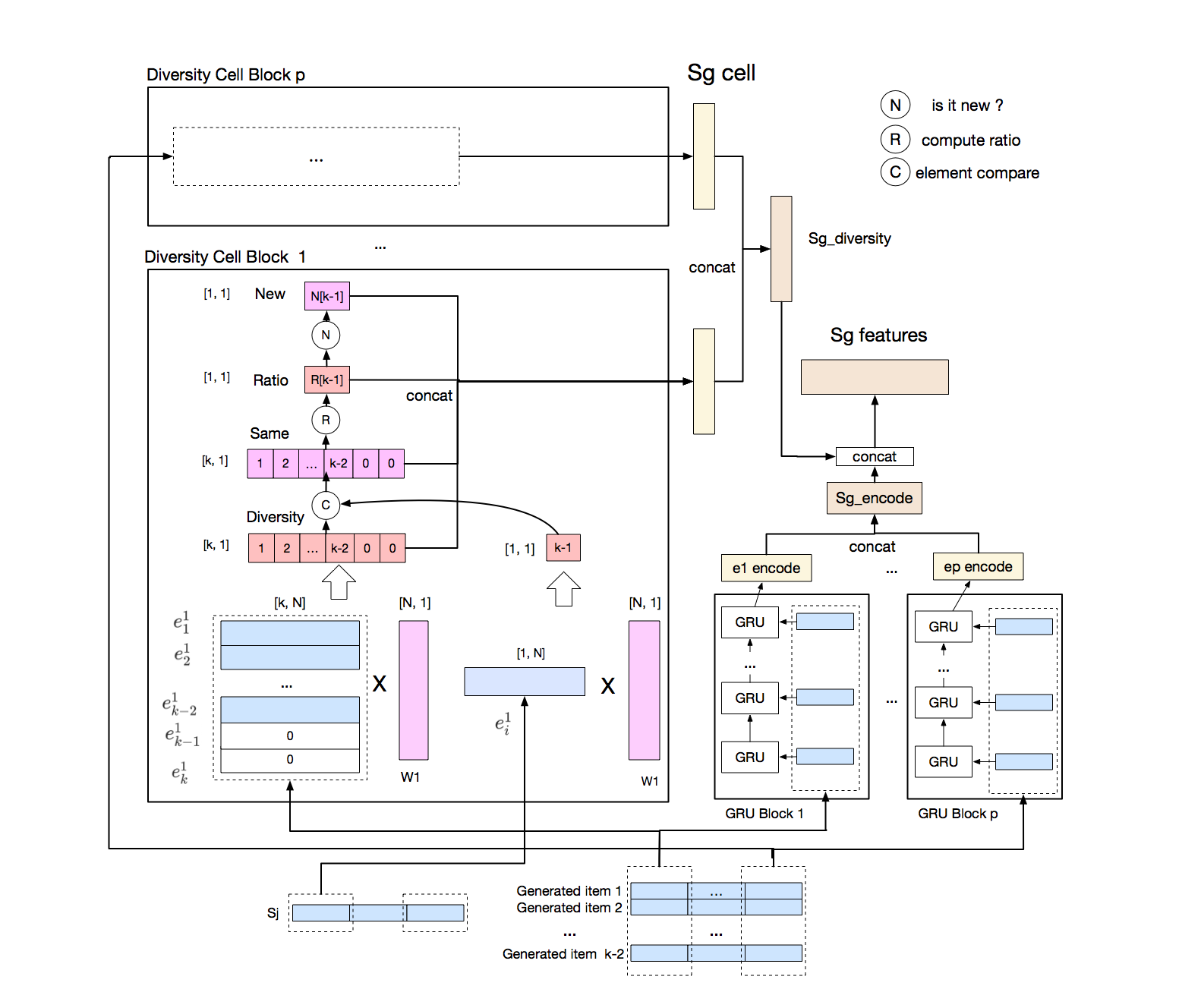}
\caption{Sg cell}
\label{fig:sgc}
\end{figure}

Now we introduce the detail of generating the information of selected items by the Sg cell. Assume that there are $p$ features of each item, such as Brand, Price and Category. The Sg cell firstly en- codes these features, and get $p$  encoding matrix, i.e., $e=(e_1,...,e_p)$. Then the encoding of the selected items can be represented as $e_{it}=(e_{ij_1},...,e_{ij_t})$. In the implementation, we choose an encoding matrix with size $k$, $E_{it}=(e_{ij_1},...,e_{ij_t},0,...,0)$. Figure \ref{fig:sgc} shows the example of the encoding in the case that there are $k-2$ items which have been selected. Besides the encoding representation, the Sg cell also compare the encoding of each candidate item $j$ with the the encoding of select items, $E_{it}$, to get the diversity information $D_{itj}$. The diversity information provides effective signals for the RL algorithm to select items with new feature values. The output of the Sg cell for each item $j$, is the combination of both the encoding information of selected items and the comparison of the diversity of the item $j$ over selected items. That is, $sg_{tj}=(E_{it},D_{itj}).$

{\bf Reward Design:} We observe that directly training the RL policy to solve the MDP fails to learn a good generation policy, as the agent can only receive the signal of the slate at the last step. As the objective function Eq.(\ref{obj}) is increasing with the conversion probability of any item, thus we choose the probability of each item as the immediate reward, $r(s_t,a_t)=p_{ia_t}.$

{\bf Efficient Exploration:} To improve the exploration ability of the RL algorithms, \cite{bellemare2016unifying} proposes the counted-based exploration methods. The counted-based exploration methods give important exploration bonus to encourage the RL algorithm to explore unseen states and actions: $r'(s_t,a_t)=r(s_t,a_t)+c*B(N(s_t,a_t))$, where $c$ is a positive constant, $N(s_t,a_t)$ counts the number of times of the pair $(s_t,a_t)$ visited by the RL policy during the training, and $B$ denotes the bonus function which is decreasing with $N(s_t,a_t)$. However, it is impossible to apply the count-based exploration method to our problem as the state space and the action space is very huge.

\cite{pathak2017curiosity} propose to learn the model that predicts the state transition function, and uses the prediction error as the exploration bonus, to motivate the policy to explore novel states that are the hardest to be predicted by the model. But the state transition in our setting is decided only by the policy, the prediction error of the model is small for any state. Thus the method can not apply to the slate re-ranking problem.

Based on the intuition that improving the diversity of items helps to explore new slates, we propose a new exploration method for the slate generation, called PPO-Exploration. At each step, we set the norm of the diversity of the picked item $a_i$ over selected items, $||D_{ita_t}||_2$ as the exploration bonus of the RL policy. That is,

\begin{equation}
\label{reward}
r'(s_t,a_t)=p_{ia_t}+c||D_{ita_t}||_2.
\end{equation}

The new reward function Eq.(\ref{reward}) trades off between the conversion probability of the picked item $a_t$ the step $t$ $p_{ia_t}$, and the degree of the diversity of the item $a_t$ over selected items, $||D_{ita_t}||_2$.
Now we propose the PPO-Exploration algorithm in Algorithm \ref{algo:ppo-exploration}, which enables an efficient exploration and exploitation trade-off.

\begin{algorithm}[H]
      \caption{The PPO-Exploration algorithm}
      \label{algo:ppo-exploration}
      \begin{algorithmic}[1]
            \STATE Initialize the actor network $\pi_{\theta}$, buffer $\mathcal{D}$
            \\
            \FOR {batch =1,...,$b$}
            \STATE Sample a batch $\{(q_1,u_1,x_1,g_{11}),...,(q_m,u_m,x_m,g_{m1})\}$ from the log data; Clear the buffer $\mathcal{D}.$\\
            \FOR {i=1,...,$m$}
            \STATE Initialize the state to be $(q_i,u_i,x_i,g_{i1}).$\\
            \FOR {t=1,...,$k$}
            \STATE Sample action $a_t$ according to the softmax distribution given by the current policy, $\pi_{\theta}(a_t=j|s_t)$\\
            \STATE Execute action $a_t$, observe new state $s_{t+1}$\\
            \STATE The Sg cell outputs the diversity bonus, $D_{ita_t}$\\
            \ENDFOR
            \STATE The critic module evaluates the generated slate $(q_i,u_i,g_{ik})$\\
            \STATE Initialize $R_t,A_t=0$\\
            \FOR {$t=k$,...,1}
            \STATE Set $R_t=r_t+\gamma*R_{t+1}$, and $A_t=R_t+c||D_{ita_t}||_2$\\
            \STATE Store transition $(s_t,a_t,r_t,s_{t+1},A_t)$ in $\mathcal{D}.$
            \ENDFOR
            \ENDFOR
            \STATE Update the policy $\pi_{\theta}$ by  minimizing the loss on the buffer $\mathcal{D}:$\\
            \STATE $L^{clip}(\theta,\theta^{'})=\frac{1}{mk}
            \sum_{s_t,a_t}min\{r_t(\theta,{\theta}^{'})A_t, clip(r_t(\theta,{\theta}^{'}), 1-\epsilon, 1+\epsilon)A_t\}$
            \ENDFOR
            \end{algorithmic}
\end{algorithm}

\section{Experiments}
Now we design experiments to validate the power of the Generator and Critic approach. We aim to answer these questions:

(1) What is the comparison result of the Full Slate Critic model with other slate evaluation methods? What is the effect of each component in the FSC model?

(2) Can the FSC model characterize the mutual influences of items?

(3) How does the PPO-Exploration algorithm compare with start of the art reinforcement learning algorithms such as PPO and Rein- force?

(4) Is the FSC model sufficient correct to be used as the evaluation for the RL policy?

(5) Can the Generator and Critic approach be applied to online e-commerce?

\subsection{Experimental Setup}
Our dataset is obtained from one of the largest e-commerce websites in the world. For the FSC model, the training data contains 100 million samples(user, query, and top-10 list). We reinforce negative samples to improve the ratio of the number of positive samples\footnote{The positive samples mean that at least one item is purchased.} over the number of negative samples to 1/50. Each sample consists of 23 features about the historical record of each 10 items, 7 ID features of each 10 items, and the real-time item features the user viewed, clicked, and purchased in one same search session(category, brand, shop, seller, etc). For each item in the slate, we set the label of the item is 1 if the item is purchased(pay) or added to cart(atc), and 0 otherwise. To tackle the problem of imbalance samples, we choose a weighted sum of the cross-entropy loss of items: pay(50), atc(4), click(1), impression(0.05). The testing dataset for the critic model contains 3 million new samples. For the generator module, the number of samples in the training dataset is 10 million, and each sample contains (user, query, top-50 candidate items). In our experiment, we set $n=50$ and $k=10$. In the implementation of Algorithm \ref{algo:ppo-exploration}, the batch size $m$ is 64, the factor of the exploration bonus $c$ is 1.

\subsection{Comparison Result of the FSC Model with Other Methods}
We compare the FSC model with the LTR method and the DLCM method in terms of the auc of three aspects: pv-pay, click-pay, and slate-pay. The pv-pay refers to purchased or not given the item impression, click-pay denotes purchased or not given the item click. Pv-pay and click-pay are both point-wise metrics. The slate-pay means at least one item of the slate is purchased or not given one impression of the whole slate. LTR(pv-pay) means that training LTR on the same dataset with the FSC model, while LTR(click-pay) denotes that training LTR on the same dataset that only contains clicked items. The labels of both two LTR models are pay behaviors. 
The DLCM model is trained on the same dataset with the FSC model except that the input of the DLCM model is 50 items rather 10 items.

\begin{table}  
\caption{Comparisons of FSC with other models}  
\begin{center}  
\begin{tabular}{|p{2.5cm}|p{1.5cm}|p{1.5cm}|p{1.5cm}|}  
\hline  
Model & pv-pay auc & click-pay auc & slate-pay auc \\ \hline  
LTR(pv-pay) & 0.861 & 0.757& 0.800 \\ \hline  
LTR(click-pay) & 0.855 & 0.785& 0.809 \\ \hline  
DLCM & 0.862 & 0.764& 0.812 \\ \hline
DNN & 0.876 & 0.790& 0.825 \\ \hline
DNN+FCN & 0.878 & 0.791& 0.826 \\ \hline
DNN+FCN+PIN & 0.884 & 0.796& 0.827 \\ \hline
DNN+FCN+PIN+Bi-GRU & 0.887 & 0.799& 0.831 \\ \hline
DNN+FCN+PIN+Bi-GRU+SAN & 0.896 & 0.813& 0.841 \\ \hline
\end{tabular}  
\end{center}
\label{table:fsc}
\end{table}

The comparison results on the testing dataset are shown in
Table \ref{table:fsc}. The last row ``DNN+FCN+PIN+Bi-GRU+SAN'' in Table 1 represents the FSC model. Results show that FSC significantly outperforms both LTR and DLCM in terms of both the slate-wise pay auc, and the point-wise item pay auc. To analyze the effect of each component of the FSC model, we also train and test its variants, from ``DNN'' to ``DNN+FCN+PIN+Bi-GRU''. ``DNN'' means that we exclude the four components discussed in Section 4.2. Note that ``DNN'' outperforms DLCM substantially, and the improvement comes from that ``DNN'' avoids the ``impressed bias'' while DLCM inputs 50 items rather than 10 real-impressed items. Comparing FSC with its variants, we claim that each component helps to improve the slate-wise pay auc. The most critical component is SAN, which shows that the FSC model successfully makes use of real-time user features and captures the real-time user behaviors.

\subsection{Visualization of the Mutual Influence of Items in the FSC Model}

\begin{figure}[H] 
\centering
\includegraphics[width=1.0\linewidth]{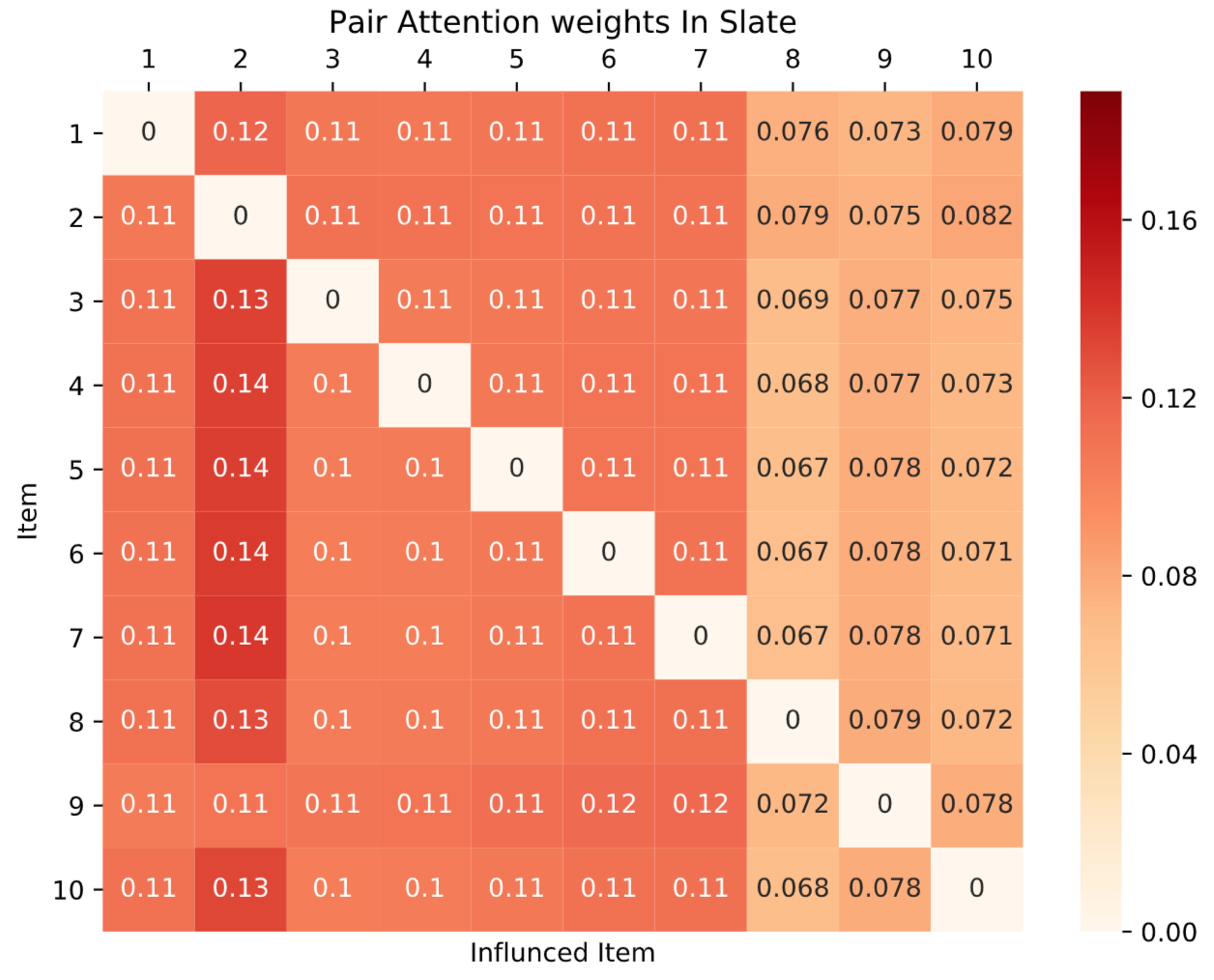}
\caption{Attention weight matrix in Pair Influence Net}
\label{fig:visual}
\end{figure}
Now we validate that the Full Slate Critic(FSC) model is able to characterize the influence of the position and the feature of other items over each item. The visualization result of the Pair Influence Net is shown in Figure \ref{fig:visual}. We input a specific slate with 10 similar items to the model and the only difference is the price: (4, 3, 5, 5, 5, 5, 5, 5, 4, 5). Then we plot the attention weight matrix as defined in Eq.(\ref{attention}). The x-axis denotes items to be influenced. We get that the weights of the 2-th column is largest among all columns, that is, the influence of other items over the 2-th item are largest. The comparison results prove that the FSC model is able to capture the effect of the distinct feature value over the user interest. It can also be observed that the weight of item 9 and item 2, 0.11 is less than the weight of item 1 and item 2, 0.12. Note that the price of item 1 and item 9 are the same, and the result can be explained by the fact that the positions of item 1 and item 2 is much closer compared with the positions of item 9 and item 2. This validates that the FSC model also carefully considers the factor of the positions of items.

\subsection{Performance Comparison of PPO-Exploration with RL Algorithms}

\begin{table}  
\caption{Comparisons results of RL algorithms}  
\begin{center}  
\begin{tabular}{|l|l|}  
\hline  
Algorithm & Replacement ratio\\ \hline  
Reinforce(no Sg cell, real reward) & 0.335 \\ \hline  
Reinforce(no Sg cell, model reward) & 0.598 \\ \hline
Reinforce(Sg cell, model reward) & 0.77 \\ \hline
PPO(Sg cell, model reward) & 0.784 \\ \hline
PPO-Exploration(Sg cell, model reward) & 0.819 \\ \hline
\end{tabular}  
\end{center}
\label{table:rl_compare}
\end{table}

We compare PPO-Exploration with start of the art RL algorithms, Reinforce and PPO. Note that the GCR framework selects the slate with a large score from two candidates: the original slate (the first 10 items from top-50 candidate items) and the re-ranked slate. Thus we use the replacement ratio to evaluate each RL algorithms, which is the frequency that the evaluation score of the slate generated by the RL algorithm is higher than that of the original slate during testing. Results are shown in Table \ref{table:rl_compare}. We get that the replacement ratio of PPO-Exploration significantly outperforms both PPO and Reinforce, which confirms that the exploration method helps to improve the performance. We also do the ablation study. Comparing row 3 with row 4 in Table 2, we validate the critical effect of the Sg cell that generates the information of selected items on the performance. Directly training the Reinforce algorithm with real rewards(the immediate reward of each item is drawn from the historical dataset) achieves the worst performance. This is because the number of historical samples is limited and model-free RL algorithms suffer from the problem of high sample complexity.

\subsection{Validation of the Effectiveness of the FSC Model on the Slate Generation}
As the FSC model is not perfect, one may doubt the correctness of using it as the evaluation of a slate generation policy. Now we validate that the FSC model can be used as the evaluation for the RL policy. We choose to apply the unbiased inverse propensity score estimator\cite{swaminathan2017off} to evaluate any RL policy $\pi_{\theta^{'}}$ with the real reward,

\begin{equation}
\label{ips}
V_{IPS}(\pi_{\theta^{'}})=\frac{1}{m}\sum_{i=1}^{m}\frac{\pi_{\theta^{'}}(L_k|s_i)}{\pi_{\theta(L_k|s_i)}r_i}.
\end{equation}

\begin{table}  
\caption{The importance sampling evaluation of RL algo- rithms}  
\begin{center}  
\begin{tabular}{|p{3cm}|l|l|}  
\hline  
Algorithm & IPS & wIPS\\ \hline  
Reinforce(no Sg cell, real reward) & $8.35*10^{-5}$ & $8.35*10^{-5}$ \\ \hline
Reinforce(Sg cell, model reward) & $2.63*10^{-4}$ & $2.48*10^{-4}$ \\ \hline
PPO-Exploration(Sg cell, model reward) & $1.4634*10^{-3}$ & $1.4636*10^{-3}$ \\ \hline
\end{tabular}  
\end{center}
\label{table:ips}
\end{table}

$r_i$ denotes the real reward of the slate, that is, the number of items purchased according to the dataset. $L_k$ is the slate generated by the policy $\pi_{\theta}$ and $\pi_{\theta}(L_k|s_i)$ is the probability that the slate $L_k$ is picked by the policy $\pi_{\theta}$. We choose ``Reinforce(no Sg cell, real reward)'' in Table \ref{table:rl_compare} as the base policy $\pi_{\theta}$ to evaluate other policies. We also take the weighted inverse propensity score(wIPS) estimator as the evaluation metric, with lower variance than IPS and asymptotically zero bias. As shown in Table \ref{table:ips}, the comparison results of RL algorithms in terms of IPS and wIPS are consistent with that of replacement ratio (evaluation results from the FSC model). That is, the FSC model is appropriate for evaluating the RL policy.

\subsection{Live Experiments}
\begin{table}  
\caption{Comparisons on the diversity of slates}  
\begin{center}  
\begin{tabular}{|p{3cm}|l|l|}  
\hline  
Bucket & brand(entropy) & price(entropy)\\ \hline  
Base & 1.342 & 1.618 \\ \hline
Test & 1.471 & 1.630 \\ \hline
\end{tabular}  
\end{center}
\label{table:diversity}
\end{table}

We apply the Generator and Critic approach to one of largest e-commerce websites in the world. The Critic and Generator module works as in Figure \ref{fig:gcr} in online experiments, and are updated weekly. During the A/B test in one month, the GCR approach improves 5.5\% number of orders, 4.3\% gmv, 2.03\% conversion rate\footnote{The main metrics of ranking algorithms in e-commerce.}, compared with the base bucket. We also compare the entropy of the brand(price) distribution of slates from the experiment bucket and the base bucket. As shown in Table \ref{table:diversity}, the GCR approach also improves the diversity of slates besides efficiency improvement.

\section{CONCLUSION}
In this paper, we propose the Generator and Critic approach to solve the main challenges in the slate re-ranking problem. For the Critic module, We present a Full Slate Critic model that avoids the ``impressed bias'' of previous slate evaluation models and outperforms other models significantly. For the Generator module, We propose a new model-based algorithm called PPO-Exploration. Results show that PPO-Exploration outperforms start of the art RL methods substantially. We apply the GCR approach to one of the largest e-commerce websites, and the A/B test result shows that the GCR approach improves both the slate efficiency and diversity.

It is promising to apply the GCR approach to generate multiple slates with different objectives, and output the final slate by some exploration methods. For the future work, it is interesting to study the combination of model-based reward output by the Critic model with model-free real reward, to improve the performance of the PPO-Exploration algorithm.

\bibliographystyle{ACM-Reference-Format}
\bibliography{sample-base}
\end{document}